\documentclass[cameraready]{Interspeech}

\usepackage{url}
\usepackage{tikz}
\usetikzlibrary{calc, arrows.meta, positioning, shapes.geometric}

\usepackage{tabularx}
\usepackage{array}
\usepackage{xcolor}
\usepackage{comment}
\usepackage{amsmath}
\usepackage{amstext}

\newif\ifshowrevisions
\showrevisionsfalse
\newenvironment{revised}{\par\begingroup}{\par\endgroup}
\newcommand{\revision}[1]{#1}

\title{Decolonizing Linguistic Policies in Automated Speech Recognition: A Framework for Cross-Culturally Competent Speech AI}

\author[affiliation={1}]{Jay L.}{Cunningham}
\author[affiliation={2}]{Mark Atta}{Mensah}
\author[affiliation={3}]{Richard}{Martinez}
\author[affiliation={1}]{João Vieira}{da Silva Neto}
\author[affiliation={1}]{Efi}{Dawodu}

\address{
  $^{1}$ DePaul University, School of Computing, RAISE Lab, USA \\
  $^{2}$ York University, Electrical Engineering and Computer Science, Canada \\
  $^{3}$ Independent Researcher, USA
}
\email{jcunni37@depaul.edu, mamensah@yorku.ca, martinezrichardme@gmail.com, jvieirad@depaul.edu, edawodu@depaul.edu}


\begin{document}

\maketitle

\begin{abstract}
This paper focuses on automatic speech recognition (ASR) and ASR-mediated voice interfaces that shape access to public services, healthcare, and education. We argue that persistent failures for low-resource, Indigenous, and non-standard language varieties are not only technical errors, but also implicit linguistic policies that reproduce colonial language hierarchies. Drawing on linguistic capital, raciolinguistic ideology, language policy research, and decolonial computing, we show how data, metrics, and model priors determine whose voices become machine-legible. We introduce the Three Harms (3M) taxonomy---Misrecognition, Misalignment, and Mistrust---and a seven-layer situatedness model for linguistic diversity in ASR and ASR-mediated voice interfaces. We then propose a participatory framework and minimum audit protocol for culturally competent ASR, positioning affected communities as co-designers, evaluators, and governance partners.
\end{abstract}

\section{Introduction}

\begin{revised}
Speech AI systems now mediate access to public services, healthcare, education, and legal processes, yet they routinely fail speakers of low-resource, Indigenous, and non-standard language varieties \cite{Radford23-RRS,Conneau21-USR}. This paper uses \textit{speech AI} as an umbrella term, but it focuses on automatic speech recognition (ASR) and ASR-mediated voice interfaces, since these systems decide whether speech becomes a transcript, an intent, or a service response. Their failures are not isolated bugs. They are predictable outcomes of how systems define which voices are legible to machines, with consequences that range from denied access to the indignity of being rendered unintelligible \cite{Koenecke20-RDI,Joshi20-TSA}. We argue that the core issue is structural, not incidental.
\end{revised}

\begin{revised}
We term these structural decisions \textbf{linguistic policies}: the design rules, assumptions, and evaluation practices that govern which language varieties a system supports, how it evaluates correctness, and what it treats as intelligible speech. The term connects our analysis to language policy and planning research, which has long examined how institutions create and maintain hierarchies among languages, speakers, and writing systems \cite{Ricento00-HAP,Spolsky19-LMS}. We use \textit{linguistic} rather than only \textit{language} to foreground speech-AI decisions that operate below and across named languages, including dialect labels, accent categories, pronunciation norms, code-switching, orthographic conventions, and error thresholds. Recent work on language technology as language management strengthens this point: technical systems do not merely reflect language policy; they can enact it through everyday use \cite{Markl25-LTLM}. In ASR, these policies operate through data, metrics, and model priors, and they reproduce colonial hierarchies of linguistic value by embedding prestige norms into the pipeline \cite{Phillipson92-LI,Thiongo86-DM}.
\end{revised}

\begin{revised}
This paper makes three integrated contributions. First, we provide a theoretical account of linguistic policies in ASR by synthesizing linguistic capital theory, raciolinguistic ideology, language policy research, and decolonial computing \cite{Bourdieu91-LAS,Rosa17-URL,Flores15-UBA,Irani10-PCL,Dourish12-RCU,Mohamed20-DSO}. This account includes a seven-layer situatedness model for linguistic diversity in speech AI, showing how language, nation, region, ethno-linguistic identity, ideology, justice, and socio-technical consequences interact. Second, we introduce the Three Harms (3M) taxonomy, Misrecognition, Misalignment, and Mistrust, together with a minimum audit protocol covering evaluator roles, sampling expectations, metrics, annotation procedures, and adjudication practices. Third, we formalize a participatory framework for culturally competent ASR that reassigns design and evaluation authority to affected language communities. We do not propose a new ASR architecture or benchmark result; rather, we contribute a human-centered evaluation framework for identifying culturally situated harms that WER-centered speech evaluation regimes under-specify.
\end{revised}

\section{Theoretical Foundations}

\begin{revised}
Our analysis integrates four intellectual traditions that illuminate the political architecture of ASR: linguistic capital theory, raciolinguistic ideology, language policy research, and decolonial computing. These traditions are not parallel lenses stacked on top of one another; together they explain how social value becomes technical behavior. Linguistic capital explains why some varieties carry institutional value. Raciolinguistic ideology explains how social judgments become listening practices. Language policy research explains how institutions manage language use. Decolonial computing explains how computational systems can extend colonial relations into technical infrastructure. Our synthesis is that ASR turns these forces into operational decisions: a supported-language menu, a dialect label, a pronunciation norm, an annotation rule, a benchmark, or a fallback response can function as policy even when no formal policy document exists.
\end{revised}

\subsection{Linguistic Capital and Market Value}

Pierre Bourdieu's concept of \textit{linguistic capital} provides a structural account of why certain language varieties are treated as more valuable than others \cite{Bourdieu91-LAS}. Language operates within a linguistic market where utterances are evaluated not only by communicative content, but also by their alignment with dominant norms. These markets are global and local. Globally, colonial and high-resource prestige varieties, such as American or British English, Metropolitan French, European Portuguese, and Peninsular Spanish, often accumulate linguistic capital that converts into social, economic, and institutional power. Locally, marginalized varieties are more often judged against the dominant standard within their own sociolinguistic ecology: African American Language against Standard American English, Afro-Brazilian Portuguese against standardized Brazilian Portuguese, Indigenous-contact Spanish varieties against dominant national or pan-Latin American Spanish norms, and regional, ethnic, or creole Englishes against locally authorized standards. Speakers of these varieties can face a symbolic tax: their speech is treated as less credible, less correct, or less worthy of institutional accommodation \cite{paffey2012language,silva2014pluricentricity,lippi2012english}.

Speech AI systems operationalize Bourdieu's linguistic market in algorithmic form. When an ASR system achieves a 5\% WER for Standard American English but a 35\% WER for African American Language \cite{Koenecke20-RDI}, or when a voice interface supports Parisian French but not Senegalese French, it does not simply reflect data availability. It reproduces the market valuation of these language varieties. The training data itself encodes market position: languages with high institutional power are overrepresented in digital corpora, while languages with lower institutional power remain ``low-resource,'' a term that can obscure the political economies of neglect that produced the resource gap \cite{Bird20-DSA}.

\begin{revised}
To avoid treating ``low-resource'' as a single technical condition, we draw on Joshi et~al.'s taxonomy of linguistic inclusion, which partitions languages by their digital and NLP resource profiles, ranging from ``Left-Behinds'' to ``Winners'' \cite{Joshi20-TSA}. We use this taxonomy descriptively, not deterministically. It shows how research attention, digital infrastructure, and institutional investment are unevenly distributed. Our claim is that ASR systems often inherit the value judgments embedded in these tiers, treating ``Winner'' languages and prestige varieties as normative and pushing culturally situated speech patterns into misrecognition, misalignment, and mistrust.
\end{revised}

\subsection{Raciolinguistic Ideologies in Technology}

\begin{revised}
Raciolinguistic ideology helps explain why speech is often judged through the perceived identity of the speaker rather than through acoustic or linguistic form alone. Rosa and Flores describe the ``white listening subject'' as a dominant perceptual stance that positions white, middle-class language practices as neutral and hears racialized speakers as deficient regardless of what they actually say \cite{Rosa17-URL,Flores15-UBA}. This framework is important for speech technology because ASR systems are not neutral listeners. They are trained, benchmarked, and deployed through social assumptions about which speech should count as ordinary.
\end{revised}

\begin{revised}
We therefore treat ASR systems as \textit{algorithmic listening subjects}. When datasets overrepresent standard varieties, pronunciation models are calibrated to prestige norms, and error tolerance thresholds are set by dominant-group performance, the system listens through a narrow institutional ear. Lawrence's analysis of Siri as a disciplining voice technology makes this point in the domain of voice assistants: users may be pressured to modify pronunciation, pacing, and linguistic style to become recognizable to the system \cite{Lawrence21-SD}. Lippi-Green's work on linguistic discrimination shows how similar judgments operate in courts, schools, and workplaces \cite{lippi2012english}. ASR extends these judgments into automated infrastructure.
\end{revised}

\subsection{Decolonial Computing}

Decolonial computing and postcolonial computing provide the fourth strand of the argument. Irani et~al.'s postcolonial computing lens shows how technology design can naturalize Western epistemologies while rendering other knowledge systems invisible or deficient \cite{Irani10-PCL}. Dourish and Mainwaring extend this critique to ubiquitous computing, arguing that systems designed for Western contexts are often deployed globally with minimal attention to local histories, infrastructures, and epistemic authority \cite{Dourish12-RCU}. Mohamed et~al. frame decolonial AI as a way to examine algorithmic exploitation, algorithmic dispossession, and the imposition of dominant epistemologies in AI systems \cite{Mohamed20-DSO}.

We build on this framework to argue that ASR represents a particularly acute site of colonial continuity. Language was a primary instrument of colonial governance \cite{Phillipson92-LI,Fanon52-BSW,Thiongo86-DM}; ASR systems that privilege colonial languages while marginalizing Indigenous and minoritized varieties can extend this governance into digital infrastructure.

\subsection{The Digital Language Divide}

These theoretical frameworks converge in what Joshi et~al. document as the digital language divide \cite{Joshi20-TSA}. Blasi et~al. show that language technology performance correlates not with linguistic complexity, but with socioeconomic power: the languages that perform worst in NLP systems are spoken by many of the world's economically marginalized populations \cite{Blasi22-SIL}. Couldry and Mejias describe a related process as data colonialism, in which data extraction reproduces older patterns of dispossession \cite{Couldry19-DCR}. For speech systems, the divide appears not only in the quantity of data, but also in whose accents, devices, domains, and interactional norms become part of benchmarked reality.

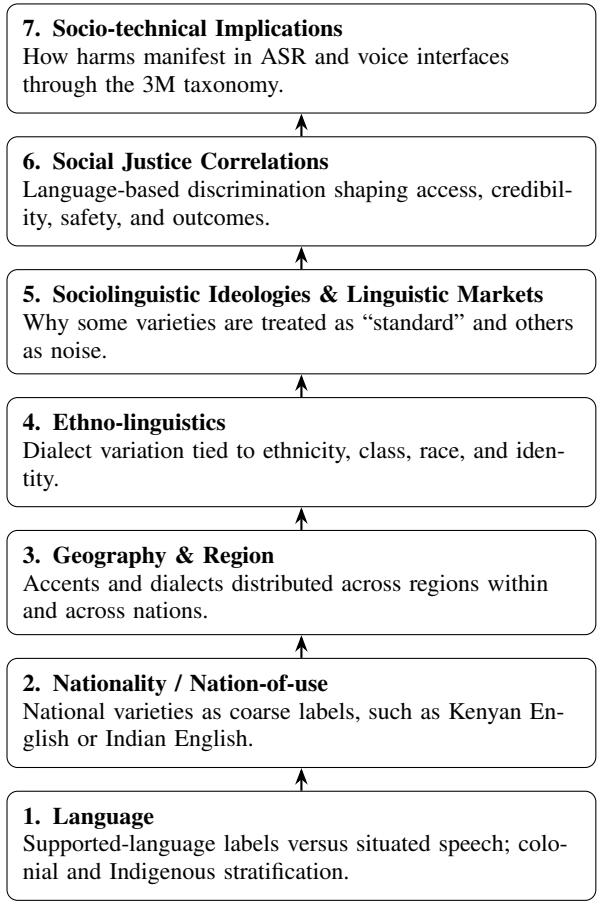
\begin{figure}[t]
\centering
\begin{tikzpicture}[
  layer/.style={draw, rounded corners, align=left, inner sep=6pt, text width=0.92\linewidth},
  arr/.style={-{Stealth[length=2.2mm]}, thick}
]
\node[layer] (l1) {\textbf{7. Socio-technical Implications}\\How harms manifest in ASR and voice interfaces through the 3M taxonomy.};
\node[layer, below=3mm of l1] (l2) {\textbf{6. Social Justice Correlations}\\Language-based discrimination shaping access, credibility, safety, and outcomes.};
\node[layer, below=3mm of l2] (l3) {\textbf{5. Sociolinguistic Ideologies \& Linguistic Markets}\\Why some varieties are treated as ``standard'' and others as noise.};
\node[layer, below=3mm of l3] (l4) {\textbf{4. Ethno-linguistics}\\Dialect variation tied to ethnicity, class, race, and identity.};
\node[layer, below=3mm of l4] (l5) {\textbf{3. Geography \& Region}\\Accents and dialects distributed across regions within and across nations.};
\node[layer, below=3mm of l5] (l6) {\textbf{2. Nationality / Nation-of-use}\\National varieties as coarse labels, such as Kenyan English or Indian English.};
\node[layer, below=3mm of l6] (l7) {\textbf{1. Language}\\Supported-language labels versus situated speech; colonial and Indigenous stratification.};

\draw[arr] (l7.north) -- (l6.south);
\draw[arr] (l6.north) -- (l5.south);
\draw[arr] (l5.north) -- (l4.south);
\draw[arr] (l4.north) -- (l3.south);
\draw[arr] (l3.north) -- (l2.south);
\draw[arr] (l2.north) -- (l1.south);
\end{tikzpicture}
\caption{A seven-layer situatedness model for linguistic diversity in ASR and ASR-mediated voice interfaces. Moving upward adds socio-technical context; colonial linguistic hierarchies shape what is treated as ``standard,'' ``supported,'' and ``valuable.''}
\label{fig:7layers}
\end{figure}

\begin{revised}
The seven-layer situatedness model in Figure~\ref{fig:7layers} is derived from this theoretical synthesis. Layer~1, language, should not be treated merely as a supported-language label. An ASR system may claim to support Portuguese, Brazilian Portuguese, or English while failing situated speech practices such as accent, rhythm, register, oral tradition, code-switching, and regional or ethnolinguistic variation. The lower layers, language, nation, and region, therefore reflect how language policy and digital resource allocation classify speakers through named languages and national markets while often ignoring situated \textit{parole}. The middle layers, ethno-linguistic identity and sociolinguistic ideology, draw from linguistic capital and raciolinguistic theory to show why dialect, accent, class, race, and perceived speaker identity shape intelligibility. The upper layers, social justice correlations and socio-technical implications, draw from decolonial computing to connect recognition errors with access, dignity, surveillance risk, and repair. We use the model as a diagnostic map: it asks what a system means by ``language,'' which layers it claims to support, which layers it ignores, and where harms become visible.
\end{revised}

\section{Positionality Statement}

In alignment with Haraway's account of situated knowledges and the privilege of partial perspective \cite{haraway2013situated} and Suchman's formulation of located accountabilities in technology production \cite{suchman2002located}, we view this paper as shaped by our standpoints as researchers whose identities, linguistic communities, and institutional locations condition what we notice as harm, what we treat as evidence, and what we imagine as responsible intervention. We do not claim neutrality. Our analysis of linguistic policies in speech AI is informed by the ways language hierarchies are lived and enforced through schooling, institutions, markets, and everyday interactions, and by our recognition that speech and language technologies can reproduce these hierarchies when they privilege prestige norms.

\begin{revised}
Our author team brings lived and scholarly experience across multiple language communities that sit in different relations to colonial and postcolonial language power, including African American English and other Afro-diasporic English varieties of North America and the Caribbean, Brazilian Portuguese and its social and ethnic variations, Nigerian English alongside Yor\`ub\'a and \`Igb\`o, Mexican and Latin American Spanish varieties, and Ghanaian English alongside Twi and Akan. This range of linguistic and cultural standpoints is central to the paper's argument that cultural competence cannot be reduced to language-level coverage or aggregate accuracy. It also informs our emphasis on harms that exceed transcription error, including pragmatic misalignment and the mistrust that follows from repeated system failure.
\end{revised}

Our standpoints also shape our methodological commitments. First, our approach prioritizes community-led and regionally grounded language technology efforts as sites of expertise rather than peripheral data sources. Second, our proposed participatory framework reflects a normative stance: affected language communities should be treated as co-designers and co-auditors of speech AI systems, with meaningful influence over evaluation criteria, deployment conditions, and pathways for repair and redress. Finally, we recognize that positionality is not resolved through disclosure alone. We treat this statement as an accountability mechanism, a reminder that our claims are partial, that our recommendations must remain responsive to community critique, and that the work ahead requires sustained collaboration beyond what any single paper can represent.

We also acknowledge that we write from university institutions in North America, which shape access to publication venues, resources, and legitimacy. This paper aims to amplify and remain accountable to Global South knowledge-making rather than treat it as supplementary.

\section{Linguistic Policies as Design Choices}

\begin{revised}
We define \textbf{linguistic policies in ASR} as the design decisions that govern which varieties a system supports, how it evaluates correctness, and what counts as intelligible speech. This definition adapts language policy theory to the speech-AI pipeline by shifting from formal language policy to computational linguistic policy. Spolsky's account of language policy as practices, beliefs, and management is useful here: ASR systems encode practices through supported-language lists, dialect labels, training-data composition, benchmark design, and pronunciation norms; beliefs through assumptions about standard speech; and management through fallback behavior, correction rules, refusal conditions, and error thresholds \cite{Spolsky19-LMS}. Markl's account of language technology as language management further supports the claim that technical artifacts can manage language use without being formal state policy \cite{Markl25-LTLM}. A speech interface therefore enacts policy when it asks accented speakers to adapt their speech, treats multilingual utterances as noise, collapses regional varieties into a single standard, or routes unsupported users into English-only fallback behavior.
\end{revised}

Table~\ref{tab:policy_signals} provides a compact diagnostic for identifying where linguistic policies are enacted in practice and how they surface in system behavior.
\begin{table}[t]
\centering
\small
\setlength{\tabcolsep}{3pt}
\renewcommand{\arraystretch}{1.1}
\caption{How to spot linguistic policies in ASR: sites of decision-making and observable signals.}
\label{tab:policy_signals}
\begin{tabularx}{\columnwidth}{p{0.28\columnwidth} X}
\hline
\textbf{Policy site} & \textbf{Concrete signals} \\
\hline
Training data & Benchmark composition skews toward prestige languages; crowdsourcing excludes rural or low-connectivity speakers. \\
Training data & Transcription guidelines enforce standardized orthography over local spellings or code-switching. \\
Metrics & Primary evaluation uses WER without domain- or community-specific weighting. \\
Metrics & Reference transcripts assume a single correct version despite multiple legitimate variants. \\
Model priors & Language model favors monolingual sequences; code-switching is penalized. \\
Model priors & Normalization rules remove honorifics or politeness markers as noise. \\
Deployment & Refusal or fallback behavior defaults to English-only responses for unsupported varieties. \\
\hline
\end{tabularx}
\end{table}
This diagnostic is not exhaustive, but it makes policy sites visible and actionable for audit and redesign.

\subsection{Training Data Curation as Policy}

The selection, sourcing, and annotation of training data constitute the most consequential linguistic policy decisions in ASR. When developers prioritize publicly available speech corpora, which overwhelmingly represent English, Mandarin, and a small number of European languages \cite{Joshi20-TSA}, they enact a policy of linguistic triage that mirrors colonial hierarchies. The term ``low-resource language'' obscures the political economy of this resource distribution: these languages are not inherently resource-poor; they have been made resource-poor through centuries of institutional marginalization \cite{Bird20-DSA}.

Consider the historical production of this scarcity. English was imposed on ethnic Africans in Ghana, Nigeria, Kenya, and South Africa, nations that each possess rich arrays of Indigenous and local languages such as Akan, Ga, Swahili, and Zulu. Spanish was imposed on Indigenous Americans and enslaved Africans in the Caribbean. Portuguese was imposed on Indigenous and Afro-Brazilian populations. French was enforced across West African nations including Senegal and Ivory Coast, and in Haiti \cite{Phillipson92-LI,Thiongo86-DM}. These colonial languages subsequently became the languages of digital infrastructure, institutional documentation, and academic publishing, which are the very sources from which training corpora are derived. When an ASR system is trained on available data, it is often trained on the documentary legacy of colonial language dominance.

\subsection{Evaluation Metrics as Policy}

Word Error Rate (WER) remains the dominant metric for ASR evaluation, yet its use as a primary measure of system quality enacts its own linguistic policy. WER treats word-level errors as equivalent, irrespective of their communicative or social consequences. It cannot capture whether a transcription error changes the meaning of an utterance, whether it renders a speaker's intent unintelligible, or whether it has differential consequences in high-stakes contexts such as healthcare or legal proceedings \cite{Markl22-LIV}.

For tonal languages, such as those found across Africa and East/Southeast Asia, WER can under-specify harm because a segmentally similar transcript may still lose meaning-bearing tonal contrasts. Yoruba, for example, uses high, mid, and low tones to encode lexical contrasts \cite{pulleyblank2004tonal,laniran2003downstep}, and tone handling is essential for Yoruba ASR because pitch can encode lexical and grammatical contrasts \cite{olusanya-2026-tone}. Chen et~al. define Tone Error Rate (TER) as a tone-aware extension that reveals errors WER collapses into a single lexical score, showing how WER can mischaracterize African-language ASR by merging phonological and tonal errors \cite{chen-etal-2026-linguistically}. Treating WER as universal therefore encodes a linguistic policy decision that privileges evaluation assumptions from dominant benchmark languages. Culturally competent ASR evaluation should supplement WER and character error rate (CER) with tone-aware measures, such as TER, that identify whether tonal substitutions introduce meaning-changing errors.

Click consonants in languages such as Xhosa, Zulu, Yeyi, Hadza, Sandawe, and Khoisan-family languages show a parallel segmental problem: a system may claim language support while deleting, substituting, or normalizing phonological features with lexical, cultural, and identity-bearing meaning. A culturally competent audit should therefore report click-specific deletion, substitution, insertion, and misclassification rates; preserve click place and manner distinctions in phoneme-level scoring; test meaning-changing minimal pairs; and include community adjudication of whether errors are minor, meaning-changing, identity-erasing, offensive, or pragmatically harmful \cite{cunningham2024understanding}. These results should accompany WER, CER, phoneme error rate, and community-weighted harm scores, since a single click deletion may matter more than an ordinary spelling error.

Moreover, WER presupposes a correct transcription, which often reflects a standardized orthography and a prestige language norm. For languages and dialects without a standardized written form, or for speakers who code-switch between multiple varieties, the concept of a single ground-truth transcription is contested \cite{poplack1980sometimes}. When ASR systems are evaluated against reference transcriptions produced by speakers of standard varieties, the evaluation metric can become a mechanism for enforcing linguistic conformity, a measurement instrument that encodes the values it claims to assess neutrally \cite{Wassink22-UEB}.

\subsection{Language Model Priors as Policy}

Language models embedded in speech systems encode assumptions about probable speech. These priors are often learned from text corpora that share the same representational biases as speech training data, then assign higher probability to utterances that conform to dominant language patterns. A Swahili speaker in Nairobi who code-switches between Swahili and English may receive lower confidence scores not because their speech is less fluent, but because the model assigns lower probability to mixed-language sequences. An Akan speaker whose speech includes tonal distinctions absent from the model's phonemic inventory may be corrected toward the nearest English approximation.

These three levels of linguistic policy, data curation, metric selection, and model priors, interact to produce a system that does not merely fail to recognize certain voices. It can also produce certain voices as unrecognizable. The decolonial insight is that exclusion is not only the absence of inclusion. It is also the presence of a normative order that constitutes some speech as unintelligible by design.

\section{The Three Harms: Misrecognition, Misalignment, and Mistrust}
\label{sec:3m}

\begin{revised}
The seven-layer model identifies the sociolinguistic conditions that ASR systems often flatten; the 3M taxonomy identifies the harms that emerge when those layers are ignored in evaluation and deployment. In this sense, the seven-layer model diagnoses where situatedness is ignored, while the 3M taxonomy diagnoses how harm appears when those layers are ignored.
\end{revised}

Conventional evaluation of ASR focuses primarily on transcription accuracy, but this captures only one dimension of harm and can mis-specify what counts as failure in multilingual, culturally situated use. In broader NLP, representational and allocational harms have been used to describe how systems reinforce stereotypes or distribute resources unevenly \cite{Blodgett20-LTP}. The 3M taxonomy translates these concerns into speech-specific, cross-cultural failure modes grounded in how people are heard, understood, and trusted in real interaction.

\begin{revised}
Table~\ref{tab:3m} summarizes the 3M taxonomy as an operational diagnostic. The table separates definitions, evidence, and harms.
\end{revised}

\begin{table*}[t]
\begingroup
\centering
\footnotesize
\setlength{\tabcolsep}{5pt}
\renewcommand{\arraystretch}{1.15}
\caption{The 3M taxonomy operationalized for ASR and ASR-mediated voice interfaces.}
\label{tab:3m}
\begin{tabularx}{\textwidth}{>{\raggedright\arraybackslash}p{0.15\textwidth} >{\raggedright\arraybackslash}p{0.27\textwidth} >{\raggedright\arraybackslash}p{0.31\textwidth} >{\raggedright\arraybackslash}X}
\hline
\textbf{Failure mode} & \textbf{Definition} & \textbf{Minimum evidence to collect} & \textbf{Likely downstream harm} \\
\hline
\textbf{Misrecognition} & The system fails to produce an accurate transcript, intent, speaker attribution, or usable output. & Word error rate (WER), Character error rate (CER); for tonal languages, Tone Error Rate (TER); refusal rate; intent error; subgroup gaps by variety, region, device, and acoustic condition. & Denied access, delayed service, exclusion from voice-mediated infrastructure, and repeated correction burden. \\
\textbf{Misalignment} & The system produces plausible output but misinterprets culturally situated meaning, such as idioms, honorifics, indirectness, or code-switching. & Community-rated pragmatic adequacy; meaning-preservation labels; code-switching, register, and implicature checks; examples of harmful normalization. & Loss of dignity, safety risks, harmful correction toward prestige norms, and erasure of locally meaningful speech practices. \\
\textbf{Mistrust} & Users disengage because the system feels unreliable, disrespectful, extractive, or surveillant. & Abandonment and opt-out rates; correction counts; trust and agency scales; complaint logs; short qualitative interviews. & Reduced use, reduced repair signals, lower adoption, and a feedback loop that widens performance gaps. \\
\hline
\end{tabularx}
\endgroup
\end{table*}

\subsection{Misrecognition}

Misrecognition encompasses the most visible failures: incorrect transcriptions, dropped words, wrong-speaker attributions, intent errors, and complete non-recognition. WER captures some of these failures, but misrecognition extends beyond word-level errors to include systematic patterns of erasure. Koenecke et~al. found large WER gaps between African American and White speakers across five major ASR systems \cite{Koenecke20-RDI}. Martin and Wright argue that ASR bias against African American Language must also be read through sociolinguistic features and institutional contexts, since biased recognition can compound discrimination in settings such as work and healthcare \cite{MartinWright23-BIA}.

For Global South language communities, misrecognition operates at an even more basic level. Speakers of Wolof, Igbo, or Haitian Creole may encounter systems that do not merely misrecognize their speech; they encounter systems that deny the existence of their language as a supported category. This categorical exclusion, often the system's refusal to attempt recognition, is a distinct form of misrecognition that WER cannot measure because there is no output to evaluate.

\subsection{Misalignment}

Misalignment occurs when a system produces a plausible transcription or response but misinterprets meaning in culturally situated speech. The words may be correct while the intended force, social positioning, or referent is wrong, which makes misalignment invisible to WER and other surface-form metrics. Misalignment matters because speech is not only lexical content. It also carries stance, politeness, authority, kinship, and local context.

Three illustrative cases make this concrete in ASR-mediated systems, where transcripts often feed downstream intent detection, translation, dialogue management, or normalization. A Twi utterance such as ``wo maame awo wo'' may function in context as admiration or praise, but a downstream literal rendering can make it appear insulting if pragmatic force is ignored. In Hindi, the contrast between ``aap'' and ``tum'' marks social distance, respect, and relational stance; a system that normalizes both into a generic second-person form, or fails to preserve the distinction in intent handling or translation, removes socially consequential meaning. A multilingual speaker may also code-switch within a single utterance, not as noise, but as a communicative strategy \cite{poplack1980sometimes}. Systems that force a single-language interpretation can change intent even when many words are transcribed correctly.

Detecting misalignment requires evaluation methods that go beyond transcription accuracy. We propose pragmatic adequacy ratings, where community evaluators judge whether the system preserved intended meaning and social stance; culturally grounded evaluation sets that include idioms, honorifics, register shifts, and locally salient frames; and implicature or intent checks that test whether the system preserves implied meaning rather than literal wording alone.

\subsection{Mistrust}

Mistrust is the experiential and relational harm that accrues when communities perceive, often reasonably, that a speech system was not designed for them. It appears in interactional signals such as abandonment, repeated corrections, opt-out behavior, and the choice to route around voice interfaces. Mengesha et~al. show that speakers of stigmatized English dialects often frame voice assistants as unintelligent, a coping strategy that still signals erosion of trust \cite{Mengesha21-IDL}. Harrington et~al. similarly show that Black older adults' experiences with a voice assistant for health information seeking were shaped by code-switching, privacy concerns, and doubts about whether the system could meet culturally specific needs \cite{Harrington22-IKC}.

Mistrust creates an adoption and repair loop that worsens inequity. As users disengage, systems receive fewer error corrections and less representative interaction data. This reduces performance for groups already underserved, which deepens mistrust. Mistrust is not simply non-use or low adoption; it is evidence of a failed legitimacy relationship between the system and the speech community. In practice, mistrust is not merely a perception problem. It is a structural mechanism that can entrench initial exclusion.

Mistrust also has a surveillance edge case. Communities with histories of state or corporate monitoring may reject speech systems even when they function accurately, especially when always-on listening, data retention, or unclear consent processes are present. Lawrence's account of voice assistants as disciplining systems helps explain why a working interface may still feel unsafe or extractive to some users \cite{Lawrence21-SD}. Cultural competence therefore requires recognizing mistrust as a rational response to risk, not as a defect in user attitudes.

The 3M taxonomy is additive, not substitutive: it complements WER and other quantitative metrics rather than replacing them. Its value lies in making visible the dimensions of harm that technical metrics alone cannot capture, and in grounding evaluation in the experiences of communities affected by speech AI's linguistic policies. \revision{Together, TER, pragmatic adequacy ratings, and trust measures show how the 3M taxonomy translates culturally situated harms into auditable outputs rather than treating them as purely theoretical categories.}

\subsection{Minimum 3M Audit Protocol}
\label{sec:minimum-3m-audit}

\begin{revised}
A minimum 3M audit requires five documented steps:
\begin{enumerate}
    \item \textbf{Define context and risk.} Researchers identify the deployment setting, target varieties, expected users, and risk level before selecting test material.
    \item \textbf{Construct a culturally grounded test set.} The test set covers ordinary service requests, domain-specific vocabulary, code-switching, idioms or indirect requests, honorific or register markers, and expected refusal cases. For a pilot audit, a small community may begin with a modest diagnostic set, but authors must report the number of speakers, utterances, minutes of audio, devices, recording conditions, and demographic or regional strata; deployment claims require larger samples and confidence intervals.
    \item \textbf{Specify evaluator roles.} A minimal panel should include fluent community evaluators for each target variety, at least one local domain expert for high-stakes settings, and one technical evaluator who computes ASR metrics.
    \item \textbf{Annotate 3M failure modes.} Annotation separates transcript error, intent error, meaning shift, register or honorific loss, unsafe normalization, refusal, and trust concern in a shared codebook, following established practice for qualitative coding and inter-annotator agreement \cite{MacQueen98-CDT,Artstein08-IAA}.
    \item \textbf{Adjudicate and define repair.} Disagreements are adjudicated by community evaluators with documented rationales rather than resolved only by external majority vote. Mistrust should be measured through behavioral indicators and calibrated trust instruments, not inferred only from system accuracy \cite{Jian00-TRM}.
\end{enumerate}
For example, in a Yoruba ASR audit, Misrecognition would not be assessed only through aggregate WER. Because Yoruba uses tone contrastively, evaluators would annotate whether the system preserves high, mid, and low tonal distinctions and whether tonal substitutions change lexical meaning. Such errors count as Misrecognition when the transcript fails to preserve a meaning-bearing contrast; they become Misalignment when a plausible output preserves surface words but changes culturally situated intent, stance, or pragmatic force. The relevant unit of harm therefore depends on the linguistic structure of the evaluated community, not on an assumption that all tonal languages require the same audit design.
\end{revised}

\section{Community-Led Responses and Speech-Relevant Efforts}

\begin{revised}
Community-led language technology work already offers alternatives to extractive development, but these efforts vary in their direct connection to speech. We separate them here into three categories: speech-specific corpora and benchmarks, NLP or machine-translation initiatives that shape language infrastructure for ASR-mediated systems, and community research networks that build local capacity. This distinction matters because an NLP project can support speech AI without itself being an ASR project.
\end{revised}

\begin{revised}
Speech-specific resources provide the closest models for culturally competent ASR development. UGSpeechData reports about 5,000 hours of validated speech across five Ghanaian languages, Akan, Ewe, Dagbani, Dagaare, and Ikposo, with distinct roles for speakers, validators, and transcribers \cite{Wiafe2025UGSpeechData}. For culturally competent ASR, UGSpeechData shows that the central constraint in low-resource settings is not only the amount of audio, but also the governance of quality, orthography, speaker recruitment, and local expertise. Recent Akan ASR benchmarking adds an empirical performance layer to this point. Mensah et~al. evaluated seven Akan ASR models across four speech domains, including culturally relevant image descriptions, informal conversations, biblical scripture readings, and spontaneous financial dialogues; the models showed domain dependence and accuracy degradation under dataset mismatch \cite{Mensah26-AkanASR}. For culturally competent ASR, the Akan case shows why evaluation must be situated by domain, register, and interactional purpose rather than inferred from a single benchmark score.

Through the 3M lens, this kind of domain mismatch is not only an accuracy problem. It can appear as misrecognition when WER degrades across domains, misalignment when culturally relevant image descriptions or financial dialogues are transcribed without preserving local pragmatic meaning, and mistrust when speakers encounter systems that work in curated benchmark settings but fail in everyday interaction.

Mozilla Data Collective (MDC), formerly Mozilla Common Voice, offers another speech-specific pathway through community-contributed recordings and validation across many languages \cite{Ardila20-CVA}. For culturally competent ASR, MDC demonstrates the value of open contribution while also showing that contribution alone does not resolve questions of community governance, consent, and downstream reuse \cite{Sloane22-PAI}.
\end{revised}

\begin{revised}
South Asian speech benchmarks also show how broad language infrastructure can become speech-specific. IndicSUPERB, developed in connection with AI4Bharat, introduced Kathbath, a labeled speech dataset with 1,684 hours across 12 Indian languages, and created benchmarks for ASR, speaker verification, speaker identification, language identification, query-by-example, and keyword spotting \cite{Javed23-IndicSUPERB}. For culturally competent ASR, IndicSUPERB shows that evaluation should include regional benchmarking and task diversity, since downstream speech harms can emerge outside aggregate transcription accuracy. AI4Bharat's broader IndicNLP infrastructure also illustrates how text corpora, benchmarks, and pretrained models can support the language-model components that sit inside ASR-mediated voice interfaces \cite{Kakwani20-INS}. Its relevance here is therefore infrastructural rather than direct ASR evidence: it shows how broader language technology ecosystems can make culturally situated speech evaluation more feasible.
\end{revised}

\begin{revised}
Some initiatives are not speech systems, but they still matter because they redistribute research authority and create language resources that ASR systems may later depend on. Masakhane, for example, is a pan-African grassroots research collective that developed participatory machine-translation resources for African languages \cite{Nekoto20-PRL}. For culturally competent ASR, Masakhane is not direct ASR evidence; its lesson is infrastructural and political, showing how locally led language technology networks can define problems from community needs, build shared resources, and treat speakers as experts rather than data sources. AmericasNLP provides a similar example for Indigenous languages of the Americas. Its speech-to-text translation competition centered Indigenous languages as direct targets of speech and translation research rather than treating them as edge cases around Spanish, English, or Portuguese \cite{Ebrahimi23-AmericasSpeech}.

For Latin American Spanish and Indigenous-contact varieties, the technical implication is that evaluation should not collapse speech into a single ``Spanish'' category. A culturally competent audit should produce regionally situated test sets and annotation guidelines that mark Indigenous-language insertions, local place names, kinship terms, discourse particles, pronunciation variants, and code-switching as meaningful linguistic evidence rather than noise. It should also produce acceptability criteria for translation or normalization: when should a system preserve a Quechua, Nahuatl, K'iche', or Mixtec term; when should it translate; and when would translation erase cultural or legal meaning? In this sense, a conceptual-to-technical bridge has the potential to become concrete: community-led language work yields test utterances, error labels, normalization rules, refusal conditions, and adjudication records that can be used to audit ASR-mediated Spanish interfaces rather than merely describe their harms \cite{Paffey2012,Ebrahimi23-AmericasSpeech}.
\end{revised}

\begin{revised}
Afro-diaspora English varieties provide a within-English comparison. English is often treated as a well-resourced language, but African American English, Jamaican Patwa, Caribbean Creoles, and other Afro-diasporic varieties remain underrepresented in training data, evaluation sets, and product-facing voice interfaces \cite{lippi2012english,rickford1999aave,patrick1999urban,holm1988pidgins}. This produces the same pattern that the 3M taxonomy captures: misrecognition through accent and dialect error, misalignment through misunderstood culturally situated expressions, and mistrust after repeated failures \cite{Koenecke20-RDI,MartinWright23-BIA,Mengesha21-IDL}. Cultural competence cannot be inferred from language-level coverage alone.
\end{revised}

Sections~2--4 show how ASR systems enact linguistic policies through data, metrics, and model priors. Section~5 operationalizes the resulting harms through the 3M taxonomy, while Section~6 shows that community-led speech and language infrastructures already offer alternatives. We now synthesize these insights into a participatory framework for culturally competent ASR and ASR-mediated voice interfaces.

\section{A Participatory Framework for Culturally Competent ASR}

Building on the theoretical analysis, positionality statement, linguistic-policy analysis, harm taxonomy, and community-led models above, we propose a participatory framework for designing and evaluating culturally competent ASR and ASR-mediated voice interfaces. The framework comprises four interconnected pillars, each addressing a different phase of the speech AI lifecycle.

\begin{revised}
The framework differs from general calls for inclusive AI in two ways. First, it operationalizes cultural competence through recurring 3M audits: systems must be assessed for misrecognition, misalignment, and mistrust. Second, it models the lifecycle as a loop of repair and redress rather than a one-time alignment exercise. Participatory auditing informs community co-design; co-design shapes deployment constraints; deployment produces feedback signals; feedback triggers further auditing and remediation. In speech settings, this loop is important because pragmatic meaning, such as honorifics, indirectness, idioms, and code-switching, can be normalized away even when transcription appears accurate.
\end{revised}

\begin{figure}[t]
\centering
\begin{tikzpicture}[
  node distance=18mm,
  box/.style={draw, rounded corners, align=center, inner sep=6pt, text width=3.0cm},
  arr/.style={-{Stealth[length=2.2mm]}, thick}
]
\node[box] (audit) {Participatory\\Auditing\\\footnotesize (3M evaluation)};
\node[box, right=of audit] (codesign) {Community\\Co-Design\\\footnotesize (needs \& norms)};
\node[box, below=of codesign] (deploy) {Equitable\\Deployment\\\footnotesize (context-fit)};
\node[box, left=of deploy] (feedback) {Feedback\\Integration\\\footnotesize (repair \& redress)};

\draw[arr] (audit) -- (codesign);
\draw[arr] (codesign) -- (deploy);
\draw[arr] (deploy) -- (feedback);
\draw[arr] (feedback) -- (audit);

\node[align=center] at ($(audit)!0.5!(deploy)$) {\small \textbf{Goal:}\\ \small Culturally competent\\ \small ASR and voice interfaces};
\end{tikzpicture}
\caption{A participatory framework for culturally competent ASR and ASR-mediated voice interfaces. The four pillars form a continuous loop; participatory auditing uses the 3M taxonomy to surface misrecognition, misalignment, and mistrust and to guide repair.}
\label{fig:4pillars}
\end{figure}
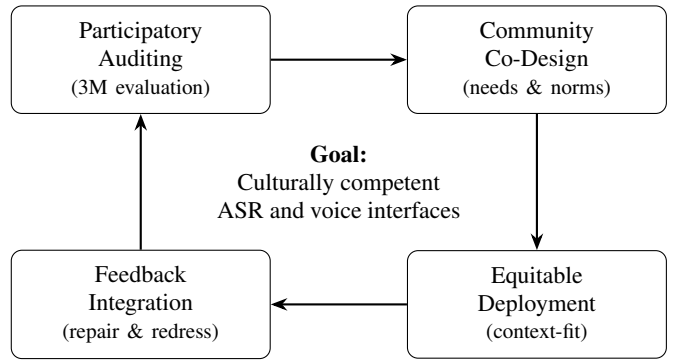

Figure~\ref{fig:4pillars} depicts the framework as a loop rather than a pipeline. The 3M taxonomy functions as the auditing instrument that links evaluation to redesign and redress.

\begin{revised}
Participation is not inherently emancipatory. Without safeguards, it can produce representational capture, community-washing, and extractive data collection without withdrawal rights or benefit-sharing governance \cite{Sloane22-PAI,Birhane22-PMP}. The framework therefore treats participation as governance, not a methodological add-on.
\end{revised}

\subsection{Pillar 1: Participatory Auditing}

Participatory auditing shifts evaluation authority from system developers to the communities the system is intended to serve. Rather than relying solely on benchmarks and quantitative metrics, it involves community members in identifying and prioritizing failures.

In relation to the 3M taxonomy, this pillar detects misrecognition, misalignment, and mistrust as distinct but interacting harms rather than treating them as undifferentiated user dissatisfaction. Its technical artifacts are a culturally situated test set, a 3M error taxonomy for the target community, annotated outputs, subgroup performance tables, and a prioritized failure list for repair.

\begin{revised}
Concretely, participatory auditing follows the minimum protocol in Section~\ref{sec:minimum-3m-audit}. Community members generate or approve test utterances reflecting real speech contexts, including code-switching, idioms, honorifics, and domain vocabulary. They evaluate outputs across the 3M dimensions: transcript or intent accuracy for Misrecognition, meaning fidelity for Misalignment, and trust or agency for Mistrust. They also prioritize which failures are most serious locally, a distinction external evaluators cannot make without situated knowledge. This approach builds on the design justice principle that those most affected should have the greatest say in evaluation \cite{CostanzaChock20-DJ}.
\end{revised}

\subsection{Pillar 2: Community Co-Design}

Community co-design involves language communities in data collection, annotation, and model development as partners from the outset. Drawing on Harrington et~al.'s deconstructed community-based design methodology \cite{Harrington19-DJA}, co-design requires three commitments: communities define accurate and appropriate behavior for their contexts; annotation guidelines reflect local linguistic norms; and communities retain governance rights over contributed speech data, including withdrawal rights and control over downstream use.

In 3M terms, co-design lets communities define what counts as ``correct,'' ``meaningful,'' and ``trustworthy'' before these criteria are converted into datasets, labels, metrics, or deployment rules. Its technical artifacts are community-authored annotation guidelines, transcript-normalization rules, pragmatic adequacy rubrics, consent templates, and acceptability criteria for system responses.

The co-design process also attends to what we call \textit{productive non-recognition}: cases in which communities may prefer that a system not recognize or record certain speech, for reasons of cultural protocol, privacy, or resistance to surveillance. This is a dimension of cultural competence that inclusion-focused frameworks often overlook: true competence includes knowing when not to listen.

\subsection{Pillar 3: Equitable Deployment}

Equitable deployment addresses governance after speech systems are built. It asks who controls the system, who benefits, and who bears the costs of failure. This pillar draws on Sambasivan et~al.'s analysis of fairness across cultural contexts \cite{Sambasivan21-RIA} and Mohamed et~al.'s decolonial AI framework \cite{Mohamed20-DSO}.

Through the 3M taxonomy, equitable deployment determines whether recognition should occur at all, under what conditions it should occur, and what safeguards are required when misrecognition, misalignment, or mistrust would create unacceptable risk. Its technical artifacts are deployment constraints, fallback and escalation rules, data-retention limits, refusal conditions, and release checklists that specify when a system should not be used.

Equitable deployment requires communities to help decide whether a system is deployed, under what conditions it operates, how failures are remediated, and how benefits from data and model improvements are distributed. This is a governance requirement, not only a technical one.

\subsection{Pillar 4: Feedback Integration}

Feedback integration closes the loop between deployment and development by channeling community-identified failures, cultural misalignments, and trust deficits back into system improvement. In practice, users need mechanisms to flag misrecognition, submit corrections, indicate meaning or intent errors, and report when the system feels disrespectful, unsafe, or not designed for them.

Within the 3M framework, feedback integration converts reported failures into repair obligations, requiring system owners to document, prioritize, and address misrecognition, misalignment, and mistrust rather than merely collecting complaints. Its technical artifacts are repair logs, correction queues, model or data update criteria, community review records, and post-remediation audit reports.

These four pillars are interdependent: participatory auditing generates the situated knowledge that informs co-design; co-design produces systems that are more amenable to equitable deployment; equitable deployment creates the conditions for sustained feedback integration; and feedback integration improves the system in ways that participatory auditing can verify.

\section{Discussion and Future Work}

\begin{revised}
This paper offers a conceptual and methodological contribution for ASR-mediated voice interfaces and speech AI. It specifies constructs, diagnostic questions, and minimum audit procedures for studying cultural competence empirically. Rather than report a new experiment, it defines what future audits should measure to evaluate harms beyond WER. The 3M taxonomy is a proposed evaluation lens, and the participatory framework is a protocol for accountable design that requires validation across communities and deployment settings.
\end{revised}

\begin{revised}
The next stage is empirical validation. We see three follow-on studies. First, a cross-linguistic ASR bias audit should evaluate commercial and open-source systems across 3--5 Global South language communities, reporting WER, tonal error rate (TER), character error rate, refusal rate, intent error, pragmatic adequacy ratings, trust measures, and subgroup gaps. Second, participatory co-design studies should test whether community-authored prompts, local annotation guidelines, and community-led adjudication change both measured performance and user trust. Third, linguistic policy analysis should map how major speech AI platforms define supported languages, unsupported varieties, reference transcripts, and fallback behavior. These studies would test whether the 3M categories are reliable across language communities, whether participatory evaluators converge or diverge from external annotators, and whether audit findings produce different design priorities than WER-only evaluation.
\end{revised}

\begin{revised}
Several limitations shape interpretation. The framework is broad, and each application requires contextual adaptation. The four-pillar structure simplifies processes that are entangled in practice. Small language communities may not support large evaluator panels; in such cases, researchers should report the constraint, narrow the claim, and preserve community authority in adjudication. The framework may inform broader speech AI, including spoken dialogue, speaker recognition, and speech synthesis, but those tasks raise additional risks around identity, biometric inference, and surveillance.
\end{revised}

We also argue for \textit{non-recognition} as a legitimate design outcome. In some contexts, opacity-by-design and selective listening are necessary for privacy, safety, and cultural protocol. Communities should determine when systems listen, what gets stored, and when speech is intentionally not recognized or retained.

\begin{revised}
Our emphasis on non-recognition is not a retreat from inclusion, but an argument for selective intelligibility or refusal, perhaps through obfuscation \cite{BruntonNissenbaum2015}: communities should determine when, how, and why their voices become machine-readable. This aligns with obfuscation and intentional friction as responses to extractive data regimes \cite{BruntonNissenbaum2015}. It also resonates with critiques of harmful classification systems, such as automatic gender recognition, where refusal can be protective and politically clarifying \cite{Keyes2018Misgendering}. We therefore position refusal as a governance commitment: defaults, interfaces, and guardrails that restrict capture, limit downstream reuse, and preserve opacity when recognition would amplify risk or reproduce colonial linguistic hierarchies \cite{CunninghamEtAl2023Solutionism}.
\end{revised}

If speech AI is to fulfill its promise of bringing people and communities closer across languages and cultures, it must reckon with the colonial legacies that have historically determined who gets to speak, who gets understood, and who gets to build the systems that listen.


\begin{revised}
\section{Generative AI Use Disclosure}
Generative AI tools were used to support editing and formatting. The authors developed the theoretical argument, framework design, source selection, and substantive analysis, and they take full responsibility for the content of the paper. No generative AI output is used as scientific evidence.
\end{revised}
\section{Acknowledgments}
This work is guided by DePaul University’s Vincentian mission \cite{University_2021}, including its commitments to human dignity, inclusion, service, and just responses to contemporary social challenges. The authors of this work see the pursuit of culturally competent ASR as aligned with these commitments insofar as it foregrounds the agency of marginalized language communities and asks how speech technologies might be designed toward repair, accountability, and equitable participation.

This work was conducted and led by the DePaul University Responsible AI Systems and Societal Experiences (RAISE) Lab, which advances research, public knowledge, and innovation on the social and ethical responsibilities of computing and human-centered technologies.

The authors would like to thank Ty Gill-Saucier (University of Washington Sociolinguistics Lab) for their feedback and trusted review. 
\bibliographystyle{IEEEtran}
\bibliography{mybib}

\end{document}